\documentclass{article}
\usepackage{spconf,amsmath,graphicx,hyperref}

\usepackage{booktabs}
\usepackage{graphicx}
\usepackage{wrapfig}
\usepackage{xpatch}
\usepackage{tabularx}
\usepackage{makecell}
\usepackage{bbding}
\usepackage{pifont}
\usepackage{xcolor}
\usepackage{xpatch}
\usepackage{multirow}
\usepackage[table]{xcolor} 
\usepackage{etoolbox}
\patchcmd{\wrapfigure}{\setlength{\intextsep}{0pt}}{}{}{}

\usepackage{framed}
\usepackage{amsmath}
\usepackage{algorithm}
\usepackage{algorithmic}

\title{Orthogonal Weight Modification Enhances Learning Scalability and Convergence Efficiency without Gradient Backpropagation}
\name{Guoqing Ma $^{1, 2,3}$  \qquad Shan Yu$^{1,2,3,\star}$ \thanks{$^{\star}$  Corresponding author: shan.yu@nlpr.ia.ac.cn} }
  
  \address{\fontsize{11}{14}\selectfont  $^{1}$ Institute of Automation, Chinese Academy of Sciences, Beijing, China \\
  \fontsize{11}{14}\selectfont $^{2}$ School of Future Technology, University of Chinese Academy of Sciences\\
      \fontsize{11}{14}\selectfont $^{3}$ Key Laboratory of Brain Cognition and Brain-inspired Intelligence Technology, Chinese Academy of Sciences}

\begin{document}
\ninept
\maketitle
\begin{abstract}
Recognizing the substantial computational cost of backpropagation (BP), non-BP methods have emerged as attractive alternatives for efficient learning on emerging neuromorphic systems. 
However, existing non-BP approaches still face critical challenges in efficiency and scalability. 
Inspired by neural representations and dynamic mechanisms in the brain, we propose a perturbation-based approach called LOw-rank Cluster Orthogonal (LOCO) weight modification. 
%
We find that low-rank is an inherent property of perturbation-based algorithms. Under this condition, the orthogonality constraint limits the variance of the node perturbation (NP) gradient estimates and enhances the convergence efficiency.
Through extensive evaluations on multiple datasets, LOCO demonstrates the capability to locally train the deepest spiking neural networks to date (more than 10 layers), while exhibiting strong continual learning ability, improved convergence efficiency, and better task performance compared to other brain-inspired non-BP algorithms. Notably, LOCO requires only O(1) parallel time complexity for weight updates, which is significantly lower than that of BP methods. This offers a promising direction for achieving high-performance, real-time, and lifelong learning on neuromorphic systems.

\end{abstract}
%

\begin{keywords}
Non-BP algorithm, Perturbation-based Optimization, Orthogonal Weight Modification
\end{keywords}
\section{Introduction}
\label{Introduction}

Neuromorphic computing has emerged as an energy-efficient alternative to traditional AI systems by leveraging brain-inspired spike-based computation~\cite{merolla2014million} and in-memory computing architectures~\cite{sebastian2020memory}. Despite its promising computational advantages, neuromorphic computing still lacks effective learning algorithms that can fully exploit its architectural parallelism and efficiency. Currently, mainstream learning algorithms can be broadly categorized into two major classes: Backpropagation (BP)-based and non-BP~\cite{lillicrap2020backpropagation}. BP requires exactly symmetric forward and backward connections to allow the gradients to be propagated back, which leads to weight transport problem~\cite{lillicrap2020backpropagation} and prevents BP from being implemented in neuromorphic systems~\cite{xiao2024hebbian}. In addition, the update-locking problem~\cite{jaderberg2017decoupled} also limits BP's parallelization, preventing it from being performed in a truly parallel manner.

Alternatively, non-BP algorithms are effectively free from the weight transport problem and make them inherently suitable for parallel computing~\cite{payeur2021burst}.
Nevertheless, many non-BP algorithms, such as Hebb-based, scalar feedback, and vector feedback~\cite{lillicrap2020backpropagation}, exhibit scalability limitations in training neural networks. While these rules demonstrate satisfactory convergence in three layers shallow networks~\cite{zhang2021self}, they often suffer from poor performance in training deeper architectures~\cite{amato2019hebbian}. Perturbation-based methods offer a potential research direction due to their rigorous mathematical foundation.
Notably, node perturbation (NP)~\cite{widrow199030} and weight perturbation (WP)~\cite{80205} are actually the zero-order gradient optimization methods, which is the mathematical essence of many non-BP algorithms~\cite{hiratani2022stability}. Therefore, enhancing NP's training efficiency is a crucial goal.
 
Nevertheless, the convergence efficiency of NP still decreases rapidly as the variance increases with the growing number of neurons in the network~\cite{hiratani2022stability}. It also loses scalability in neural networks with more than five layers. 
The state of art, non-BP training algorithms so far can train SNNs no more than 5 layers, e.g., SBP can train 2- to 4-layer networks~\cite{zhang2021self}, and SoftHebb can train 5-layer networks~\cite{moraitis2022softhebb}.
In contrast, the brain possesses hundreds of billions of neurons while maintaining remarkable efficiency. Therefore, understanding the remarkable learning scalability of brain remains an open question critical for neuroscience and for designing efficient AI systems that avoid BP's limitations in large-scale neuromorphic computing.

Recent neuroscience findings have revealed two significant phenomena about learning efficiency, including (1) that neural representations that need to be separated are often orthogonal to each other~\cite{flesch2022orthogonal}, suggesting that synaptic modifications may be aimed at minimizing interference between different tasks 
\cite{zeng2019continual}; and (2) brain dynamics are often manifested in low-dimensional manifolds~\cite{goudar2023schema}, suggesting that learning happens in a low-rank parameter space. However, existing non-BP algorithms for Spiking Neural Networks (SNNs) have not investigated the impact of orthogonality and low-rank properties on scalability and convergence efficiency.

In this work, we propose LOw-rank Cluster Orthogonal (LOCO) weight modification algorithm. Compared with other non-BP algorithms, LOCO has better convergence and scalability, which opens a promising avenue for training neuromorphic systems. Specifically, our contributions are as follows:

\begin{itemize}

    \item We demonstrate that low-rank is an inherent property of perturbation-based algorithms. Under this condition, the orthogonality constraint limits the variance of the NP gradient estimates and enhances the convergence efficiency.

    
    \item Our results demonstrate that LOCO is capable of training a spiking neural network with more than 10 layers, whereas previous non-BP works were limited to a maximum of 5 layers, highlighting the scalability of our approach. 
    
    \item With higher convergence efficiency, LOCO achieves state-of-the-art performance on several datasets and exhibits the ability to overcome catastrophic forgetting. 

    
\end{itemize}



\begin{figure*}
  \centering
  
  \includegraphics[width=.9\textwidth]{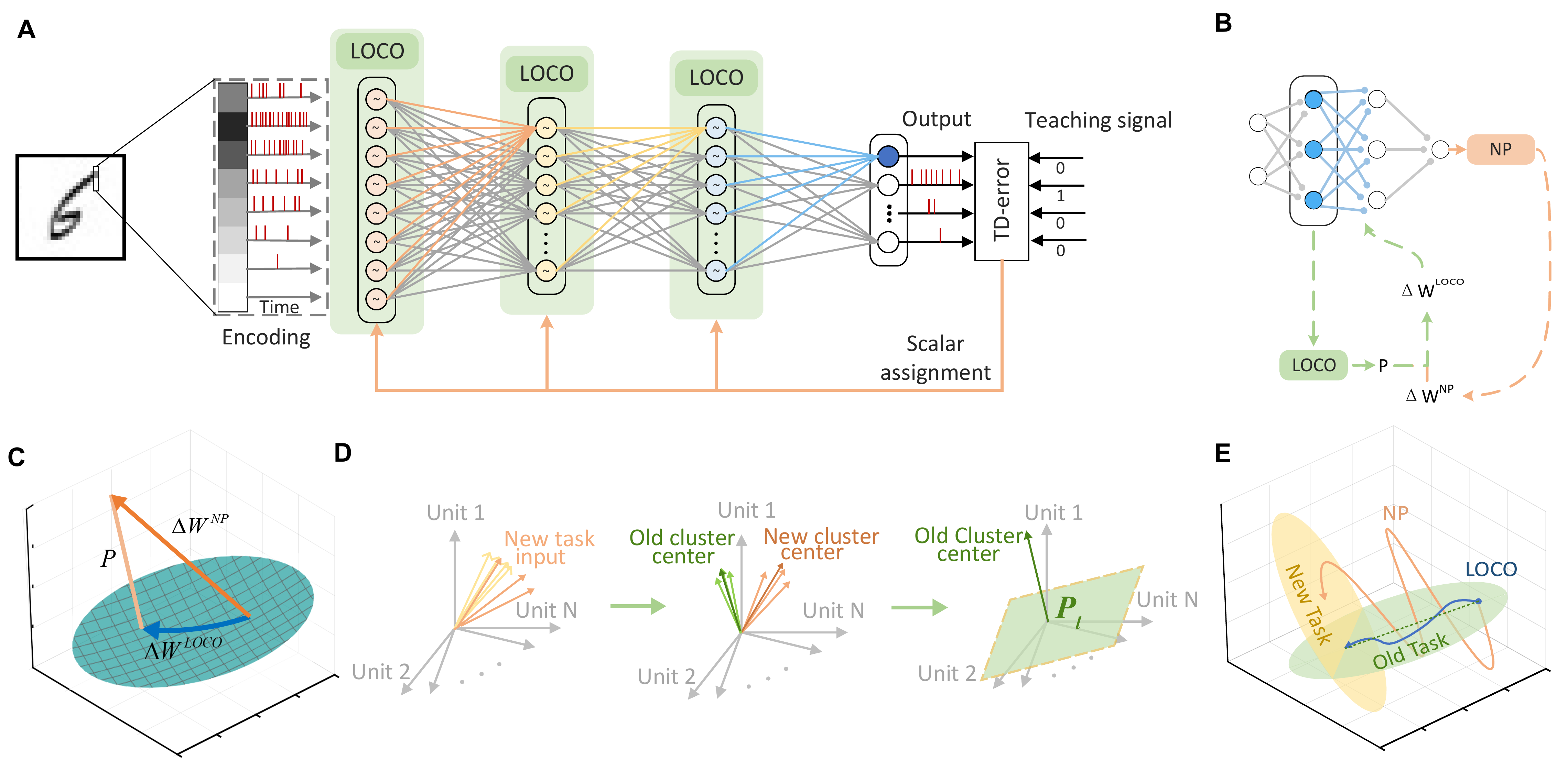}
  \vspace{-15pt} 
  \caption{\textbf{Schematic diagram of LOCO. (A)} The multi-layer architecture of SNN. The weights connecting to the same postsynaptic neurons are marked with the same color, representing the unit of weight modification in LOCO (Eq.~\ref{main-1}). \textbf{(B)} The weight modification with NP and LOCO. \textbf{(C)} During the training of new tasks, the NP weight modification $\Delta {{\rm{W}}^{NP}}$, is projected to the subspace (green surface), in which good performance for old tasks can be maintained. As a result, the actually implemented weight modification is $\Delta {{\rm{W}}^{LOCO}}$. \textbf{(D)} The process of calculating the projection matrix $P_l$. \textbf{(E)} The reduced variation during the parameter searching process between NP and LOCO. NP searches in a high-dimensional space (orange line) with high variance. LOCO reduces the variance and searches in a low-dimensional subspace (green planes).}
  \label{fig:intro}
    \vspace{-15pt} 
\end{figure*}

 
 

\vspace{-10pt} 
\section{Related work}
\vspace{-5pt} 

\noindent\textbf{Training method for deep SNN.} Non-BP rules are effective only for networks with relatively few layers, e.g., no more than 5 layers. For instance, the Hebbian learning rule~\cite{millidge2022predictive, Zhang2025EfficientRL} is effective in shallow networks, faces challenges in maintaining stability and learning efficiency in deeper networks~\cite{amato2019hebbian}. In a recent study, the Spiking Backpropagation (SBP) algorithm, which extended the depth of optimizable layers in spiking neural networks from 2 to 4 in SNN~\cite{zhang2021self}. SoftHebb, which allows effective training for up to five hidden layers~\cite{moraitis2022softhebb}. Although vector feed back learning rules, like Feedback Alignment (FA)~\cite{lillicrap2016random}, Direct Feedback Alignment (DFA)~\cite{nokland2016direct}, have been applied to train networks with multiple linear hidden layers~\cite{frenkel2019learning, DBLP:journals/jr/MaWYZ22}, as multiple linear hidden layers are equivalent to a single linear layer, the networks trained in this way cannot achieve the same capability as deep nonlinear networks~\cite{lillicrap2016random}.



\noindent\textbf{ANN-to-SNN conversion method.} A more hardware-oriented approach to achieve higher accuracy in SNN is to convert an offline trained deep artificial neural network (ANN) to a neuromorphic spiking platform (ANN-to-SNN conversion)~\cite{stanojevic2024high}. These methods first train an ANN and then transfer the weights to an SNN~\cite{neil2016learning}. Using this approach, models have been developed that obtained excellent accuracy performance. 
However, these methods cannot address the issue of online training for neuromorphic chips.


\noindent\textbf{NP and orthogonality.} NP is widely used as a non-BP algorithm, due to four properties~\cite{hiratani2022stability}: 1) It is applicable to a broad variety of networks and tasks. 2)  It has a rigorous mathematical convergence foundation. In the NP algorithm, synaptic weights are updated according to how the error changes when a small perturbation is added to each node of a neural network~\cite{widrow199030}. If a perturbation improves performance, the weights are modified so that the perturbation is consolidated, and vice versa. Orthogonal weight modification (OWM) is often used in continual learning tasks~\cite{zeng2019continual, ma2025mitigating} and has also been applied in SNN~\cite{xiao2024hebbian}. However, these methods are designed for continual learning tasks and are theoretically unsuitable for retraining previously learned tasks. In contrast, our method uses clustering-based orthogonality, which allows for flexible training of both new and previously learned tasks while maintaining stability. 



\vspace{-10pt} 
\section{Method} \label{Method}
\vspace{-5pt} 

\subsection{Pipeline Overview}

We deploy a multi-layer architecture composed of Leaky Integrate and Fire (LIF) neurons, as illustrated in Fig.~\ref{fig:intro}A, with the number of layers varying from three to eleven. The learning process involves two forward propagation runs. The first run is the standard one

\begin{equation}\label{ec-6}
{s^l}(t) = {f_l}\left( {{W^l}{s^{l - 1}}(t)} \right),l = 1,2,...,L,
\end{equation}

\noindent where ${s^0}(t)$ and ${s^L}(t)$ are the input spike and the output spike respectively, and ${f_l}( \cdot )$ represents the propagation dynamics of the LIF neurons, which is an element-wise function. 
The second run introduces a random perturbation ${\sigma {\xi ^l}}$ to the membrane potential of individual neurons of the hidden layer

\begin{equation}\label{ec-7}
{{\tilde s}^l}(t) = {f_l}\left( {{W^l}{{\tilde s}^{l - 1}}(t) + \sigma {\xi ^l}} \right),l = 1,2,...,L.
\end{equation}

The perturbations cause a minor difference in the network output and the loss compared to the first run. We name the difference in the loss between the two runs as the temporal difference (TD) error, which is broadcast to the hidden layer of the network to guide learning. Specifically, based on the average firing rates of neurons in each layer and the TD error, the weight modification ${\Delta W_l^{{\text{NP}}}}$ is obtained using the NP algorithm (Fig.~\ref{fig:intro}B). 

\begin{equation}\label{ec-8}
\Delta W_l^{{\text{NP}}} =  - \frac{\eta }{\sigma }(\sigma {\xi ^l})(\tilde \ell ({{\tilde s}^L},{s^0}) - \ell ({s^L},{s^0})){{{\bf{x}}}_{l - 1}}^T,
\end{equation}

\noindent where $\tilde \ell ({{\tilde s}^L},{s^0}) - \ell ({s^L},{s^0})$ is TD error. ${{\bf{x}}}_{l - 1}$ represents the fire rate of the $l$ layer input spike.

\subsection{Low-rank Cluster Orthogonal Weight Modification (LOCO)}
The gradient variance estimated by NP is high~\cite{hiratani2022stability}. To reduce variance, we introduce an orthogonal constraint (Fig.~\ref{fig:intro}C). The orthogonal constraint can preserve important dimensions while excluding the unimportant ones via projecting

\begin{equation}\label{main-1}
\Delta W{_l^{{LOCO{\text{ }}}^T}} = {P_l}\Delta W{_l^{{{\text{NP }}}^T}},
\end{equation}

\noindent where ${P_l}$ is an orthogonal projection operator. ${\Delta W_l^{{\text{NP}}}}$ represents the weight modification vector obtained by the node perturbation algorithm.

However, orthogonal constraints can only compress a limited number of dimensions through projection. The number of compressed dimensions is generally equal to the number of classes, ranging approximately from 0 to 10.  

The dimension 's' represents the number of principal components in matrix $A$ (Eq.~\ref{ec-15}). It starts at 0 ($A$ is initially empty) and increases with training, up to 10 for the 10-class MNIST task.

\begin{figure}[ht]
    \centering
    \includegraphics[width =.8\linewidth]{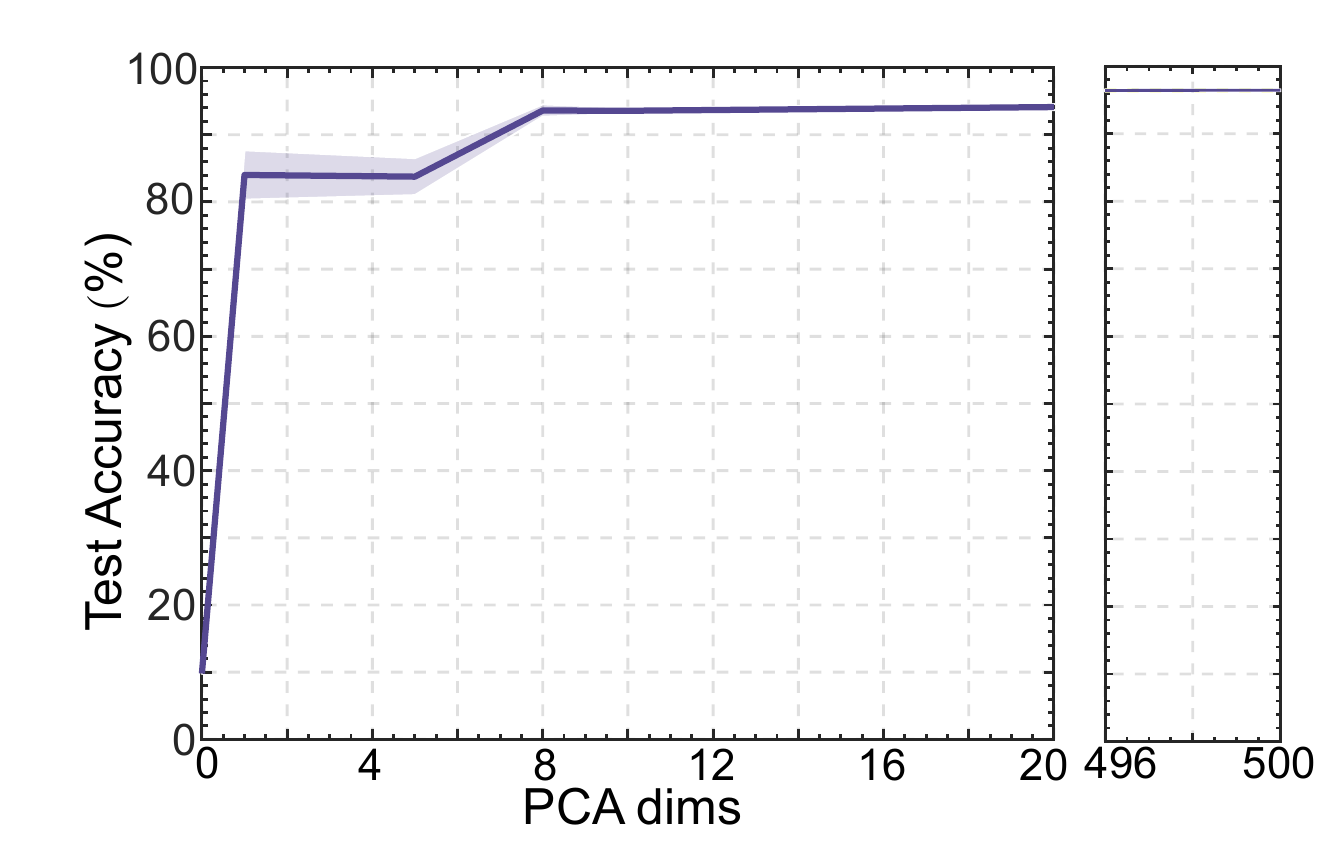}
    \vspace{-15pt}
    \caption{\textbf{Low-rank is an inherent property of perturbation-based algorithm.}}
    \label{rank}
    \vspace{-5pt}
\end{figure}

Even so, we found that orthogonal constraints still significantly improve the convergence efficiency. 
To explain this phenomenon, the experiment (Fig.~\ref{rank}) demonstrates that when the weight modification is constrained to a low-rank space, the neural network's accuracy does not significantly decrease, indicating that the weight modification resides in a low-dimensional space.  
Therefore, we propose that the efficiency improvement may be due to the overlap between the dimensions compressed by the orthogonal constraint and the low-rank space where the weight modification resides.  
Therefore, the upper bounds for the learning rates can be expressed as:

\begin{equation}\label{main-1-2}
{\eta _{LOCO}} = \gamma{\eta _{NP}},
\end{equation}

\noindent where $\eta _{LOCO}$ and $\eta _{NP}$ denote the maximum learning rates for ensuring convergence in LOCO and NP respectively. The scaling factor $\gamma  = \frac{r}{r - o} > 1$, where $r$ is the inherent rank of weight modification. $r-o$ is the overlapped dimensions between the space compressed by the orthogonal constraint and the low-rank space of weight modification. Because it ensures $\gamma>1$, LOCO maintains higher convergence efficiency compared to NP. Notably, as $r$ decreases, $\gamma$ increases, indicating that low-rank property can further enhance the convergence efficiency. 
Due to space limitations, the proof cannot be completed here. If needed, we can provide it during the review stage.

Intuitively, LOCO reduces the search space compared to NP, reducing training variance and task interference, thus accelerating convergence (Fig.~\ref{fig:intro}E).


\subsection{Calculation of projection matrix} \label{Calculation of projection matrix}


Previously proposed orthogonal algorithms are designed for continual learning problems to overcome catastrophic forgetting. The continual learning task paradigm was strictly set to learn different tasks sequentially. Previous orthogonal algorithms were unable to relearn or adjust old tasks.  
To address this issue, we propose clustered orthogonal weight modification (Fig.~\ref{fig:intro}D). 

%
The core idea is to reduce the variance of node perturbations via a projection matrix $P_l$. Since $P_l$ is rank-deficient, it eliminates perturbation components in certain directions, thereby reducing the noise space in gradient estimation. To minimize the impact on existing knowledge, we adopt orthogonal projection to preserve the original parameter space as much as possible~\cite{11227331}.

By retaining some of the input and clustering, we dynamically compute the projection matrix $P_l$.
The $P_l^{}$ is computed as follows

\begin{equation}\label{ec-15}
P_l^{} = I - {A_l}{(A_l^T{A_l})^{ - 1}}{A_l}^T,
\end{equation}

\noindent where  ${A_l} \in {{\mathcal R}^{n \times s}}$ represents the principal components of all categories except the current one. 
$n$ denotes the number of neurons in the hidden layer and also represents the dimensionality of the space in which the input resides. 
$s$ is the number of principal components to be preserved, equivalent to the number of cluster centers $c$ minus one. ${A_l}$ is calculated by 

\begin{equation}\label{ec-16}
\begin{array}{l}{U_{l - 1}} = \mathrm {kmeans}({X_{l - 1}},c)\\
{\bf{u}}_{l - 1}^j = \mathrm {nearest}({\bf{x}}_{l - 1},{U_{l - 1}})\\
{A_l} = \{ {U_{l - 1}}/{\bf{u}}_{l - 1}^j\} \end{array},
\end{equation}

\noindent where ${X_{l - 1}} = [{\bf{x}}_{l - 1}^0,{\bf{x}}_{l - 1}^1,...,{\bf{x}}_{l - 1}^B] \in {{ R}^{n \times B}}$ is a matrix randomly selected from input. $B$ is the buffer size. ${\bf{x}}_{l - 1}$ denotes the input fire rate in layer $l$. The k-means clustering algorithm is employed to cluster $B$ directions into $c$ cluster centers, represented by ${U_{l - 1}} = [{\bf{u}}_{l - 1}^0,{\bf{u}}_{l - 1}^1,...,{\bf{u}}_{l - 1}^{\text{c}}]$. 
The cluster center ${\bf{u}}_{l - 1}^j$ nearest to current input ${\bf{x}}_{l - 1}$ in ${U_{l - 1}}$  is found in terms of angular distance, i.e., the $j$-th cluster center. By removing the $j$-th cluster center from all cluster centers, the final preserved principal component ${A_l}$ are obtained. 

\begin{figure}[ht]
    \centering
    \includegraphics[width =1.\linewidth]{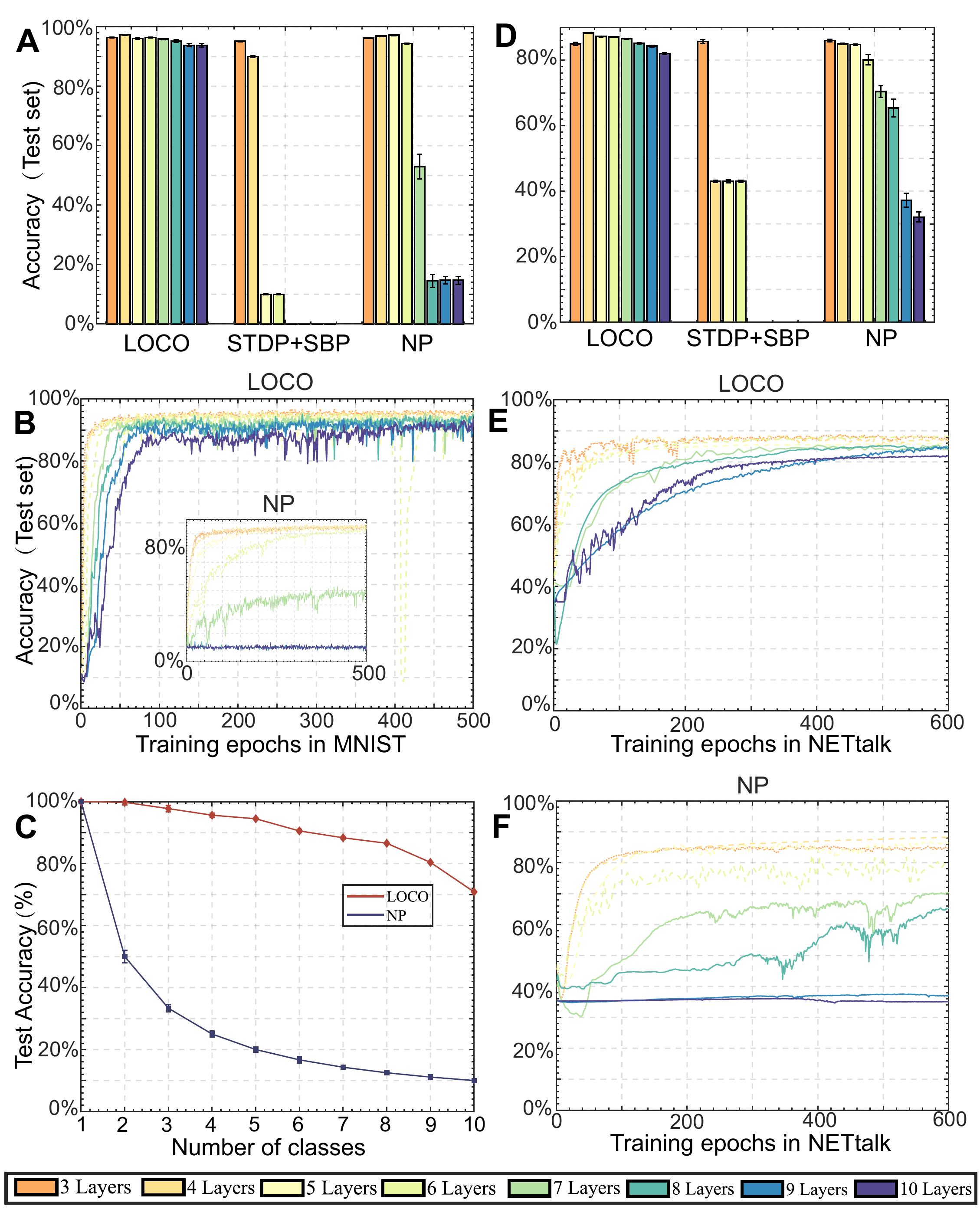}
    \vspace{-15pt}
    \caption{\textbf{LOCO demonstrates superior scalability, enhanced convergence efficiency, and continual learning capabilities without gradient backpropagation.}  \textbf{(A to C)} Performance on the MNIST dataset. \textbf{(A)} LOCO can train a deeper spiking neural network than STDP+SBP and NP. Performance of network with differrent hidden layers are color-coded. \textbf{(B)} Learning dynamics of LOCO for neural networks ranging from 3 to 10 layers. 
    The inset shows the accuracy curve of NP. \textbf{(C)} Accuracy curves for continual learning task on MNIST. The horizontal axis represents the current class being learned, with each class trained sequentially. The vertical axis indicates the average classification accuracy on all previously learned classes.
    \textbf{(D to F)} Performance in the phonetic transcription task, presented in the same manner as that in (A) and (B). The shaded regions and error bars represent the variance of the performance across five runs with different seeds.
}
    \label{10layers}
    \vspace{-15pt}
\end{figure}

\begin{figure}[ht]
    \centering
    \includegraphics[width =.65\linewidth]{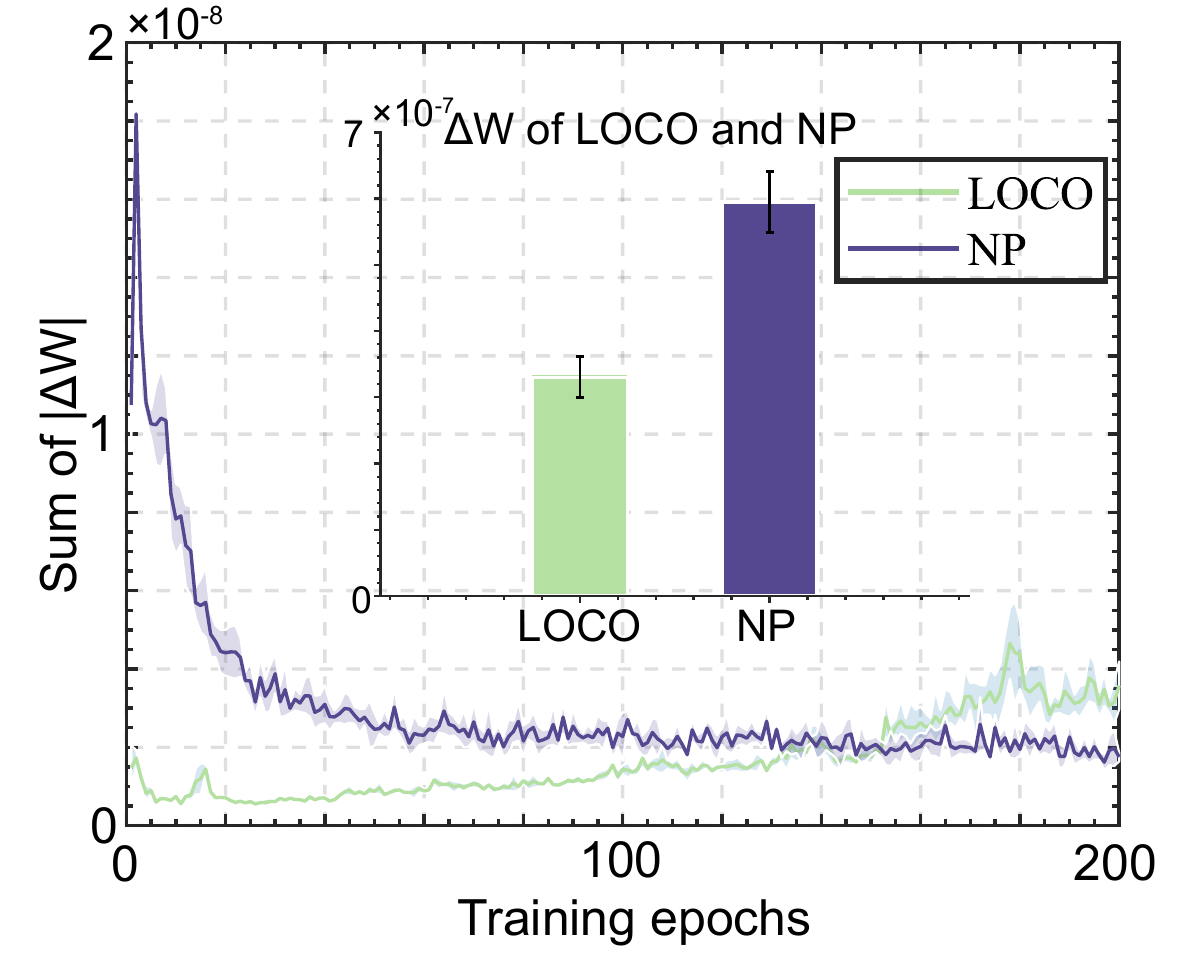}
    \vspace{-15pt}
    \caption{\textbf{The magnitude of weight changes in LOCO is smaller than that of NP. This implies that LOCO is more stable and energy-efficient.} 
}
    \label{deltaW}
    \vspace{-15pt}
\end{figure}

\vspace{-5pt} 
\section{Experiments}
\vspace{-5pt} 
\subsection{Experimental Setup}
\vspace{-5pt} 
To verify the effects of the LOCO algorithm in training SNNs, we use four different learning tasks:  (1) recognition of handwritten digits using the Modified National Institute of Standards and Technology (MNIST) dataset~\cite{yann1998minist}; and (2) phonetic transcription using the NETtalk dataset~\cite{sejnowski1987parallel}; (3) Imagenette~\cite{Howard_Imagenette_2019} a subset of dataset from Imagenet. 

\vspace{-10pt} 
\subsection{Results and Analyses} 
\textbf{LOCO improves the scalability and convergence efficiency.}
For learning hand digit recognition on MNIST dataset, we use SNNs comprising 784 input neurons, 500 neurons per hidden layer, and 10 output neurons.
The network depth in our experiments varies from 3 to 11 layers. We compare the accuracy achieved by LOCO, NP, and STDP+SBP~\cite{zhang2021self} algorithms across different network depths. 
Experimental results show that LOCO can train a 10-layer neural network without significant performance loss. In contrast, NP can only train a 5-layer network, and STDP+SBP can only train a 4-layer network. Exceeding these layer counts results in significant performance degradation (Fig.~\ref{10layers}A).
In addition, we find that the accuracy of LOCO-trained SNNs improves faster and reaches a higher plateau compared to NP (Fig.~\ref{10layers}B). 


To examine the performance of LOCO in complex tasks, we test its capability of phonetic transcription on the NETtalk dataset, in which the network needs to classify 116 classes using 26 output neurons. 
We find that the LOCO algorithm exhibits advantages that are very similar to those of the MNIST task. Specifically, compared to the NP method, it could train deeper  networks with 10 layers (Fig.~\ref{10layers}D), leads to faster learning (Fig.~\ref{10layers}E, F), and achieves better accuracy.

To further verify LOCO's capability in handling more challenging scenarios, we also compared it against non-BP methods, including SoftHebb, FA, and DFA, on the Imagenette benchmark dataset. 
LOCO maintains its ability to converge in an 11-layer convolutional spiking neural network, while also attaining superior accuracy. 

\noindent\textbf{Continual learning capability.}
Beyond that, we also test the continual learning ability of the LOCO on MNIST (Fig.~\ref{10layers}C). In a continual learning task, the network sequentially learns to recognize different sets of handwritten digits. 
The results show that the NP algorithm suffers from catastrophic forgetting, as the performance rapidly declines as more new classes are learned. In contrast, LOCO exhibits stronger continual learning capability, showing much less interference among different tasks. 

\noindent\textbf{LOCO learning in the low rank space.}
To verify that low-rank is an inherent property of perturbation-based algorithms, we restrict weight modifications to a low-dimensional subspace. The low-dimensional subspace is spanned by the first $k$ input principal components using PCA.
The results (Fig.~\ref{rank}) indicate that there is little impact on accuracy when the weight modification space is constrained to no less than 8.

\noindent\textbf{LOCO is more stable.} 
Fig.~\ref{deltaW}C shows that the magnitude of the weight changes in LOCO is smaller than that of NP. The smaller magnitudes of weight adjustments imply that computational hardware can achieve a reduction in energy consumption, and the LOCO is more stable than the NP.

\vspace{-10pt} 
\section{Conclusion} \label{Conclusions and limitations}
\vspace{-5pt} 

Non-BP algorithms offer a promising direction for optimizing neuromorphic systems~\cite{zhang2023edge}. Previous works cannot guarantee convergence or can only guarantee convergence in shallow networks, e.g., no more than 5 layers~\cite{ zhang2021self}, and linear networks~\cite{lillicrap2016random}. Perturbation-based algorithms exhibit scalability limitations due to their high variance. In this paper, we demonstrate that LOCO efficiently trains a spiking neural network with more than 10 layers.
Our work shows that perturbation algorithms possess an inherent low-rank property. Based on this, the orthogonality constraint significantly reduces the variance of gradient estimates and ultimately enhances the convergence efficiency. 
For the limit of scalability, we note that the performance typically begins to degrade beyond 15-20 layers under the current setup. In future work, we plan to incorporate techniques such as batch normalization and residual connections to enable the training of deeper SNNs, which represents a key direction for our ongoing research.  
LOCO offers a path for reliable and efficient brain-inspired learning. It demonstrates that simple scalar error feedback, combined with appropriate weight constraints, enables efficient training of complex networks, making non-BP rules like LOCO highly suitable for neuromorphic computing.

\vspace{-10pt} 
\section{Acknowledgment} \label{Acknowledgment}
\vspace{-5pt} 
This work was supported in part by the Strategic Priority Research Program of the Chinese Academy of Sciences (CAS)(XDB1010302), CAS Project for Young Scientists in Basic Research, Grant No. YSBR-041 and Beijing Natural Science Foundation Haidian Original Innovation Joint Fund Project (L232035, L222154).




\vfill\pagebreak

\bibliographystyle{IEEEbib}
\bibliography{strings,refs}

\end{document}